\documentclass{article}

\usepackage{arxiv}

\usepackage[utf8]{inputenc} %
\usepackage[T1]{fontenc}    %
\usepackage{hyperref}       %
\usepackage{url}            %
\usepackage{booktabs}       %
\usepackage{amsfonts}       %
\usepackage{nicefrac}       %
\usepackage{microtype}      %
\usepackage{graphicx}
\usepackage{caption}
\usepackage{subcaption}
\usepackage{amsmath}
\usepackage{siunitx}
\usepackage{appendix}
\usepackage{amssymb}
\usepackage{multirow}
\usepackage{xcolor}
\usepackage{textcomp}
\usepackage{hyperref}

\usepackage[bottom,symbol*]{footmisc}

\DeclareMathOperator{\clip}{clip}
\DeclareMathOperator{\round}{round}
\DeclareMathOperator{\quantize}{quantize}
\DeclareMathOperator{\dequantize}{dequantize}

\title{Integer Quantization for Deep Learning Inference: Principles and Empirical Evaluation}

\author{
Hao Wu$^1$ \and Patrick Judd$^1$ \and Xiaojie Zhang$^1$ \and Mikhail Isaev$^2$\thanks{Work done during an internship at NVIDIA} \and Paulius Micikevicius$^1$ \and \\
\begin{tabular}{cc} 
$^1$NVIDIA & $^2$Georgia Institute of Technology \\
 \texttt{\{skyw, pjudd, viczhang, pauliusm\}@nvidia.com} & \texttt{michael.v.isaev@gatech.edu} 
\end{tabular}
}

\begin{document}
\maketitle

\renewcommand*{\thefootnote}{\arabic{footnote}}
\setcounter{footnote}{0}

\begin{abstract}
Quantization techniques can reduce the size of Deep Neural Networks and improve inference latency and throughput by taking advantage of high throughput integer instructions. 
In this paper we review the mathematical aspects of quantization parameters and evaluate their choices on a wide range of neural network models for different application domains, including vision, speech, and language. 
We focus on quantization techniques that are amenable to acceleration by processors with high-throughput integer math pipelines. We also present a workflow for 8-bit quantization that is able to maintain accuracy within 1\% of the floating-point baseline on all networks studied, including models that are more difficult to quantize, such as MobileNets and BERT-large.
\end{abstract}

\section{Introduction}

 While 32-bit single-precision floating-point was the dominant numerical format for Deep Learning (DL) applications, more recently a variety of alternative formats have been proposed to increase the computational performance of deep learning applications. It is becoming commonplace to train neural networks in 16-bit floating-point formats, either IEEE fp16~\cite{MP_training2017} or bfloat16~\cite{TPU_imagenet}, supported by most DL accelerators. 
 Once trained, neural networks can be deployed for inference using even lower-precision formats, including floating-point, fixed-point, and integer. Low-precision formats offer several performance benefits. First, many processors provide higher throughput math pipelines the low-bit formats, which can speed up math-intensive operations, such as convolutions and matrix multiplications. Second, smaller word sizes reduce memory bandwidth pressure, improving performance for bandwidth-limited computations. Third, smaller word sizes lead to lower memory size requirements, which can improve cache utilization as well as other aspects of memory-system operation.

\begin{table}[b]
\centering
\begin{tabular}{llrr}
\toprule
Input Data type & Accumulation Data type & Math Throughput & Bandwidth Reduction \\
\midrule
FP32 &                  FP32 &              1x &                  1x \\
FP16 &                  FP16 &              8x &                  2x \\
INT8 &                 INT32 &             16x &                  4x \\
INT4 &                 INT32 &             32x &                  8x \\
INT1 &                 INT32 &            128x &                 32x \\
\bottomrule
\end{tabular}
\caption{Benefits of lower precision data types for tensor operations on the NVIDIA Turing GPU architecture}
\label{tab:turing_speedup}
\end{table}

In this paper we focus on integer quantization for neural network inference, where trained networks are modified to use integer weights and activations so that integer math pipelines can be used for many operations. 
Table~\ref{tab:turing_speedup} lists the relative tensor operation throughputs of various data types on the NVIDIA Turing Graphics Processing Unit (GPU) architecture~\cite{nvidia2018turing}. 
Math-intensive tensor operations executed on 8-bit integer types can see up to a 16x speed-up compared to the same operations in fp32.
Memory-limited operations could see up to a 4x speed-up compared to the fp32 version, due to the smaller word size. 
Other processors, such as TPUv1~\cite{TPUv1}, Intel CPUs with VNNI instructions~\cite{Intel_VNNI}, and a number of emerging accelerator designs also provide significant acceleration for int8 operations. 
The process of neural network quantization can be automated by software tools~\cite{TrtQuantGtc, Intel_distiller} or controlled manually. In either case, care must be taken to minimize any impact quantization has on the model accuracy.

In this paper we review the mathematical fundamentals underlying various integer quantization choices (Section~\ref{sec:fundamentals}) as well as techniques for recovering accuracy lost due to quantization (Section~\ref{sec:acc-recovery}). Section~\ref{sec:workflow} combines this information into a recommended workflow. In Section~\ref{sec:experiments} and the Appendices we present empirical evaluation of various quantization choices on a wide range of network models from different application domains - image processing, language modeling, language translation, and speech recognition. These models include the major network topologies - convolutional networks, recurrent networks, as well as attention-based networks. With the presented workflow for int8 quantization we are able to maintain model accuracy within 1\% of each baseline floating-point network, even for the networks that are known to be challenging to quantize, such as MobileNets and BERT-large.

\section{Related Work}

Vanhoucke et al.~\cite{vanhoucke2011improving} showed that earlier neural networks could be quantized after training to use int8 instructions on Intel CPUs while maintaining the accuracy of the floating-point model.
More recently it has been shown that some modern networks require training to maintain accuracy when quantized for int8.
Jacob et al.~\cite{jacob2018quantization} described models optimized for inference where all inference operations were performed with integer data types. 
Here batch normalization layers were folded into the preceding convolution layer before quantization, reducing the number of layers that needed to be executed during inference.
Krishnamoorthi~\cite{krishnamoorthi2018quantizing} evaluated various quantization methods and bit-widths on a variety of Convolutional Neural Networks (CNNs).
He showed that even with per-channel quantization, networks like MobileNet do not reach baseline accuracy with int8 Post Training Quantization (PTQ) and require Quantization Aware Training (QAT).
McKinstry et al.~\cite{mckinstry2018discovering} demonstrated that many ImageNet CNNs can be finetuned for just one epoch after quantizing to int8 and reach baseline accuracy. They emphasized the importance of using an annealing learning rate schedule and a very small final learning rate.
They also set the quantization range based on a percentile of activations sampled from the training set.
Instead of using fixed ranges, Choi et al.~\cite{choi2018pact} proposed PACT which learns the activation ranges during training.

Much of the earlier research in this area focused on very low bit quantization~\cite{courbariaux2014training,han2015deep,zhou2016dorefa}, all the way down to ternary (2-bit)~\cite{zhu2016trained,mellempudi2017ternary} and binary weights~\cite{courbariaux2015binary} and activations~\cite{rastegari2016xnor,hubara2016binarized}.
These works showed that for lower bit-widths, training with quantization was required to achieve high accuracy, though accuracy was still lower than the floating-point network on harder tasks such as ImageNet image classification~\cite{ILSVRC15}.
They also demonstrated the importance of techniques such as using higher precision for weight updates and the Straight-through Estimator (STE) for gradient backpropagation~\cite{bengio2013estimating}.
Also, in many cases the first and last layer were not quantized, or quantized with a higher bit-width, as they are more sensitive to quantization~\cite{zhou2016dorefa,rastegari2016xnor,hubara2016binarized}.
Multi-bit quantization schemes use either uniform~\cite{courbariaux2014training,zhou2016dorefa}, or non-uniform quantization~\cite{han2015deep,zhu2016trained,mellempudi2017ternary,baskin2018uniq}.
Uniform quantization enables the use of integer or fixed-point math pipelines, allowing computation to be performed in the quantized domain.
Non-uniform quantization requires dequantization, e.g. a codebook lookup, before doing computation in higher precision, limiting its benefits to model compression and bandwidth reduction.
This paper focuses on leveraging quantization to accelerate computation, so we will restrict our focus to uniform quantization schemes.

While much of the aforementioned work has focused on CNNs for image classification, there are also many examples of applying quantization to other types of network architectures.
Wu et al.~\cite{wu2016google} described how Google's Neural Machine Translation (GNMT), which employs a Long Short Term Memory (LSTM) Recurrent Neural Network (RNN), was trained with hard range constraints on multiple tensors to be more amenable to PTQ. A similar strategy was taken on MobileNet v2~\cite{sandler2018mobilenetv2}, which restricts activations to be in the range [0, 6] (ReLU6).  
Bhandare et al.~\cite{bhandare2019efficient} quantized the smaller base Transformer~\cite{Transformer2017} model targeting the int8 VNNI instructions on Intel CPUs. They use KL-Divergence~\cite{TrtQuantGtc} to calibrate the quantization ranges and apply PTQ.
Zafrir et al.~\cite{zafrir2019q8bert} quantized BERT~\cite{devlin2018bert} to int8 using both PTQ and QAT.
In this paper, we present an evaluation of int8 quantization on all of the major network architectures with both PTQ and QAT.

More complex methods have also been proposed for training quantized models.
Distillation has been used to train a quantized ``student'' model with a high precision, and often larger, ``teacher'' model.
It has been applied to training quantized CNNs~\cite{mishra2017apprentice,polino2018model}, LSTMs~\cite{polino2018model} and Transformers~\cite{junczys2018marian}.
Leng et al.~\cite{leng2018extremely} used the Alternating Direction Method of Multipliers (ADMM) as an alternative to STE when training quantized model.
These methods generally target lower bit-width quantization, as QAT has been shown to be sufficient for int8 quantization. We have also found QAT to be sufficient for int8 quantization on the models we evaluated, and as such we chose not to included these methods in our evaluation of int8 quantization.

\vspace{0pt}
\section{Quantization Fundamentals}
\label{sec:fundamentals}

We focus on uniform integer quantization as it enables computing matrix multiplications and convolutions in the integer domain, allowing the use of high throughput integer math pipelines. Uniform quantization can be divided in to two steps. First, choose the range of the real numbers to be quantized, clamping the values outside this range. Second, map the real values to integers representable by the bit-width of the quantized representation (round each mapped real value to the closest integer value).

In this Section we will consider higher precision floating-point formats like fp16 and fp32 to be real numbers for the purpose of discussion.
Enabling integer operations in a pre-trained floating-point neural network requires two fundamental operations:

\texttt{Quantize}: convert a real number to a quantized integer representation (e.g.~from fp32 to int8).

\texttt{Dequantize}: convert a number from quantized integer representation to a real number (e.g.~ from int32 to fp16).

We will first define the quantize and dequantize operations in Section~\ref{sec:range_map} and discuss their implications in neural network quantization in Sections~\ref{sec:granularity} and~\ref{sec:comp_cost}. 
Then we will discuss how the real ranges are chosen in Section~\ref{sec:calib}.

\subsection{Range Mapping} 
\label{sec:range_map}

Let  $[\beta, \alpha]$ be the range of representable real values chosen for quantization and $b$ be the bit-width of the signed integer representation. 
Uniform quantization transforms the input value $x \in [\beta, \alpha]$ to lie within $[-2^{b-1}, 2^{b-1}-1]$, where inputs outside the range are clipped to the nearest bound. 
Since we are considering only uniform transformations, there are only two choices for the transformation function: \(f(x) = s \cdot x + z\) and its special case \(f(x) = s \cdot x\), where $x, s, z \in \mathbb{R}$. In this paper we refer to these two choices as \textit{affine} and \textit{scale}, respectively.

\setlength{\abovecaptionskip}{20pt plus 3pt minus 2pt} %

\begin{figure}[t]
\centering
\begin{subfigure}{0.49\textwidth}
\includegraphics[width=\textwidth]{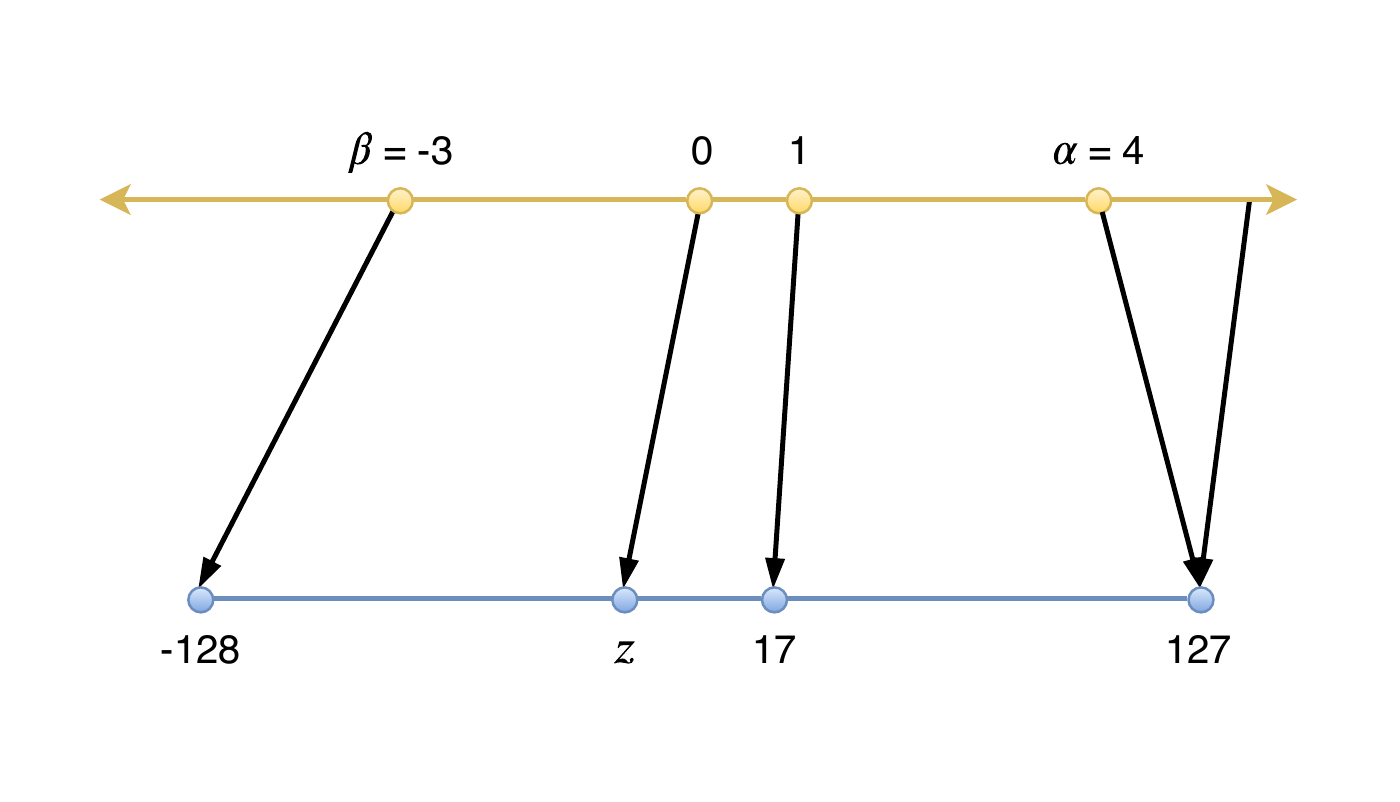}
\caption{Affine quantization}
\label{fig:mapping-affine}
\end{subfigure}%
\begin{subfigure}{0.49\textwidth}
\includegraphics[width=\textwidth]{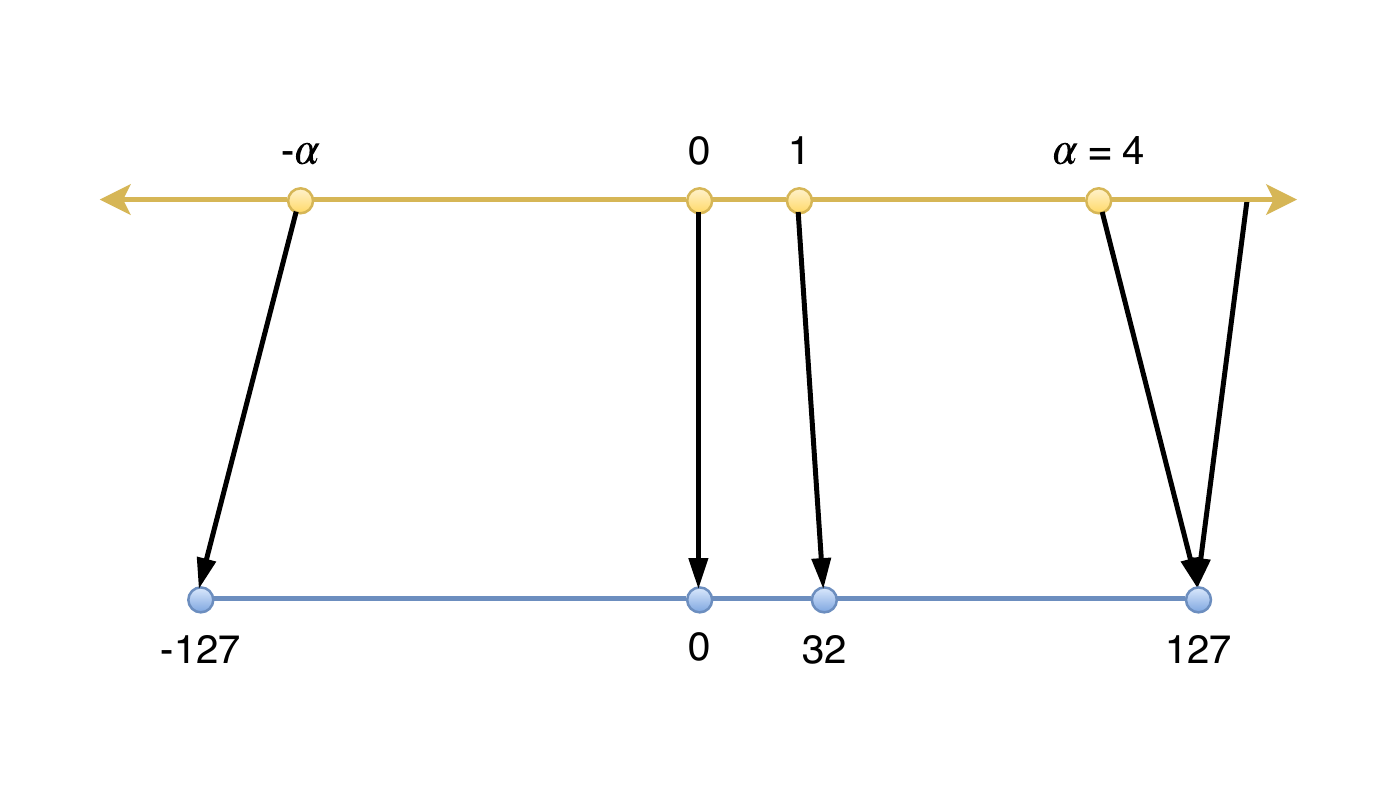}
\caption{Scale quantization}
\label{fig:mapping-scale}
\end{subfigure}
\caption{Quantization mapping of real values to int8}
\label{fig:mapping}
\end{figure}

\subsubsection{Affine Quantization}

Affine quantization maps a real value $x \in \mathbb{R}$ to a $b$-bit signed integer  $x_q \in \{-2^{b-1}, -2^{b-1}+1,\ldots, 2^{b-1}-1\}$.
Equations~\ref{eq:affine_scale} and~\ref{eq:affine_zero} define affine transformation function, \(f(x) = s \cdot x + z\):

\vspace{5pt}

\begin{equation}
s =\frac{2^{b}-1}{\alpha - \beta}
\label{eq:affine_scale}
\end{equation}

\begin{equation}
z =-\round(\beta \cdot s)-2^{b-1} 
\label{eq:affine_zero}
\end{equation}

\vspace{5pt}

where $s$ is the scale factor and $z$ is the zero-point - the integer value to which the real value zero is mapped.
In the 8-bit case, $s=\frac{255}{\alpha - \beta}$ and $z=-round(\beta \cdot s) - 128$. 
Note that $z$ is rounded to an integer value so that the real value of zero is exactly representable. 
This will result in a slight adjustment to the real representable range $[\beta, \alpha]$~\cite{jacob2018quantization}.

\newpage
The quantize operation is defined by Equation~\ref{eq:clip} and~\ref{eq:affine_quant}:

\vspace{5pt}

\begin{equation}
\clip (x, l, u)
\begin{cases}
l, ~~x < l \\ 
x, ~~l \le x \le u \\ 
u, ~~x > u \\
\end{cases}
\label{eq:clip}
\end{equation}

\begin{equation}
x_q = \quantize (x, b, s, z)  = \clip(\round(s \cdot x + z), -2^{b-1}, 2^{b-1}-1)
\label{eq:affine_quant}
\end{equation}
\vspace{5pt}

where \(\round()\) rounds to the nearest integer.
Figure~\ref{fig:mapping-affine} shows the mapping of real values to int8 representation with affine quantization. Note that $s$ is the ratio of the integer-representable and chosen real ranges.

Equation~\ref{eq:dequant} shows the corresponding dequantize function, which computes an approximation of the original real valued input, $\hat{x} \approx x$.

\vspace{5pt}

\begin{equation}
\hat{x} = \dequantize(x_q, s, z)  = \frac{1}{s}(x_q - z) %
\label{eq:dequant}
\end{equation}

\vspace{5pt}

\subsubsection{Scale Quantization}

Scale quantization performs range mapping with only a scale transformation. 
For simplicity we describe the symmetric variant of scale quantization (often called symmetric quantization~\cite{krishnamoorthi2018quantizing}), where the input range and integer range are symmetric around zero. 
This means that for int8 we use the integer range $[-127, 127]$, opting not to use the value -128 in favor of symmetry. 
For 8-bit quantization, losing one out of 256 representable values is insignificant, but for lower bit quantization the trade-off between representable values and symmetry should be re-evaluated. 

Figure~\ref{fig:mapping-scale} illustrates the mapping of real values to int8 with scale quantization. Equation~\ref{eq:scale_scale} and~\ref{eq:scale_quant} define scale quantization of a real value $x$, with a chosen representable range $[-\alpha, \alpha]$, producing a $b$-bit integer value, $x_q$.

\vspace{5pt}

\begin{equation}
s =\frac{2^{b-1}-1}{\alpha}
\label{eq:scale_scale}
\end{equation}
\begin{equation}
x_q = \quantize (x, b, s)  = \clip(\round(s \cdot x ), -2^{b-1}+1, 2^{b-1}-1)
\label{eq:scale_quant}
\end{equation}

\vspace{5pt}
Equation~\ref{eq:dequant-scale} shows the corresponding dequantize operation for scale quantization.

\vspace{5pt}

\begin{equation}
\hat{x} = \dequantize(x_q, s) = \frac{1}{s} x_q
\label{eq:dequant-scale}
\end{equation}

\vspace{5pt}

\subsection{Tensor Quantization Granularity}
\label{sec:granularity}

There are several choices for sharing quantization parameters among tensor elements. 
We refer to this choice as quantization granularity. At the coarsest, per-tensor granularity, the same quantization parameters are shared by all elements in the tensor. The finest granularity would have individual quantization parameters per element. 
Intermediate granularities reuse parameters over various dimensions of the tensor - per row or per column for 2D matrices, per channel for 3D (image-like) tensors, etc.

We will consider two factors when choosing granularity: impact on model accuracy and computational cost. To understand the computational cost, we will examine matrix multiplication (note that this results in no loss of generality for math-intensive operations since convolutions can be expressed as matrix multiplications~\cite{chetlur2014cudnn, im2col}). 

\newpage
Consider a linear (fully-connected) layer that performs a matrix multiplication $Y=XW$, where 
$X = (x_{ik}) \in \mathbb{R}^{m\times p}$ is the input activation tensor, 
$W = (w_{kj}) \in \mathbb{R}^{p\times n}$ is the weight tensor, and
$Y = (y_{ij}) \in \mathbb{R}^{m\times n}$ is the output tensor.
The result of the real-valued matrix multiplication $Y = XW$ can be approximated with quantized tensors $X_q = (x_{q,ik}) \in \mathbb{Z}^{m\times p}$ and $W_q = (w_{q,kj}) \in \mathbb{Z}^{p\times n}$, by first dequantizing them, and then performing the matrix multiplication. First, consider tensors quantized at the finest granularity, per-element, with scale quantization:

\begin{equation}
y_{ij} = \sum_{k=1}^{p}{x_{ik} \cdot w_{kj}} \approx 
\sum_{k=1}^{p}{\dequantize(x_{q,ik}, s_{q, ik}) \cdot \dequantize(w_{q,kj}, s_{w,kj})} = \sum_{k=1}^{p}{\frac{1}{s_{x,ik}}x_{q,ik} \cdot \frac{1}{s_{w,kj}}w_{q,kj} }
\label{eq:dot-product-granularity}
\end{equation}

In order to use integer matrix multiplication the scales must be factored out of the summation on the right-hand side of Equation~\ref{eq:dot-product-granularity}, for which the scales must be independent of $k$:
\begin{equation}\begin{split}
\frac{1}{s_{x,i} \cdot s_{w,j}}\sum_{k=1}^{p}{ x_{q,ik} \cdot w_{q,kj}}
\label{eq:scale_only_quant}
\end{split}\end{equation}

Thus, integer matrix multiplication is possible as long as the quantization granularity is per-row or per-tensor for activations and per-column or per-tensor for weights.
For activations, only per-tensor quantization is practical for performance reasons. In the above formulation different rows belong to either different batch instances or items in a sequence and thus row count can vary at inference time.
This prevents the per-row scaling factor from being computation offline (which would not be meaningful for different instances in a mini-batch), whereas determining them online imposes a compute overhead and in some cases results in poor accuracy (Dynamic Quantization discussion in~\cite{zafrir2019q8bert}).

For maximum performance, activations should use per-tensor quantization granularity. Weights should be quantized at either per-tensor or per-column granularity for linear layers
of the form $Y=XW$ (per-row for linear layers of the form $Y=XW^T$).
The corresponding granularity to per-column in convolutions is per-kernel, or equivalently per-output-channel since each kernel produces a separate output channel~\cite{krizhevsky2012imagenet,lecun1990handwritten}.
This is commonly referred to as ``per-channel'' weight quantization in literature and we follow that convention~\cite{jain2019trained,kozlov2020neural,krishnamoorthi2018quantizing,nagel2019data,rusci2019memory}.
We examine the granularity impact on accuracy in Section~\ref{sec:wt_only_quant}.

\subsection{Computational Cost of Affine Quantization}
\label{sec:comp_cost}

While both affine and scale quantization enable the use of integer arithmetic, affine quantization leads to more computationally expensive inference.
As shown in equation~\ref{eq:scale_only_quant}, scale quantization results in an integer matrix multiply, followed by a point-wise floating-point multiplication. Given that a typical dot-product in a DNN comprises 100s to 1000s of multiply-add operations, a single floating-point operation at the end is a negligible cost. Furthermore, if per-tensor quantization is used for both arguments, a single floating-point multiplier is needed and is part of the GEMM API (often referred to as alpha) in BLAS libraries~\cite{dongarra1990set}.

Affine quantization yields a more complex expression:
\begin{equation}
\begin{split}
y_{ij} & \approx \sum_{k=1}^{p}{
      \frac{1}{s_x}(x_{q,ik}-z_x)\frac{1}{s_{w,j}}(w_{q,kj}-z_{w,j})}   \\
    & = \frac{1}{s_xs_{w,j}} \bigg(
      \underset{(1)}{\sum_{k=1}^{p}{x_{q,ik} w_{q,kj}}}
    - \underset{(2)}{\sum_{k=1}^{p}{(w_{q,kj} z_x + z_x z_{w,j})}}
    - \underset{(3)}{\sum_{k=1}^{p}{x_{q,ik} z_{w,j}}}
    \bigg)
\end{split}
\label{eq:affine_matmul}
\end{equation}

Computation can be broken down into three terms, as annotated in Equation~\ref{eq:affine_matmul}.
The first term is the integer dot product, just as in scale quantization (Equation~\ref{eq:scale_only_quant}).
The second term consists of only integer weights and zero-points. 
As a result, this term can be computed offline, only adding an element-wise addition at inference time. If the layer has a bias then this term can be folded in without increasing inference cost.
The third term, however, involves the quantized input matrix $X_q$, and thus cannot be computed offline. 
This extra computation, depending on implementation, can introduce considerable overhead, reducing or even eliminating the throughput advantage that integer math pipelines have over reduced precision floating-point. 
Note that this extra computation is incurred only if affine quantization is used for the weight matrix. Thus, to maximize inference performance we recommend using scale quantization for weights. 
While affine quantization could be used for activations without a performance penalty, we show in later sections that scale quantization is sufficient for int8 quantization of all the networks we studied.

\newpage
\subsection{Calibration}
\label{sec:calib}

\begin{figure}[tb]
\centering
\includegraphics[scale=0.35]{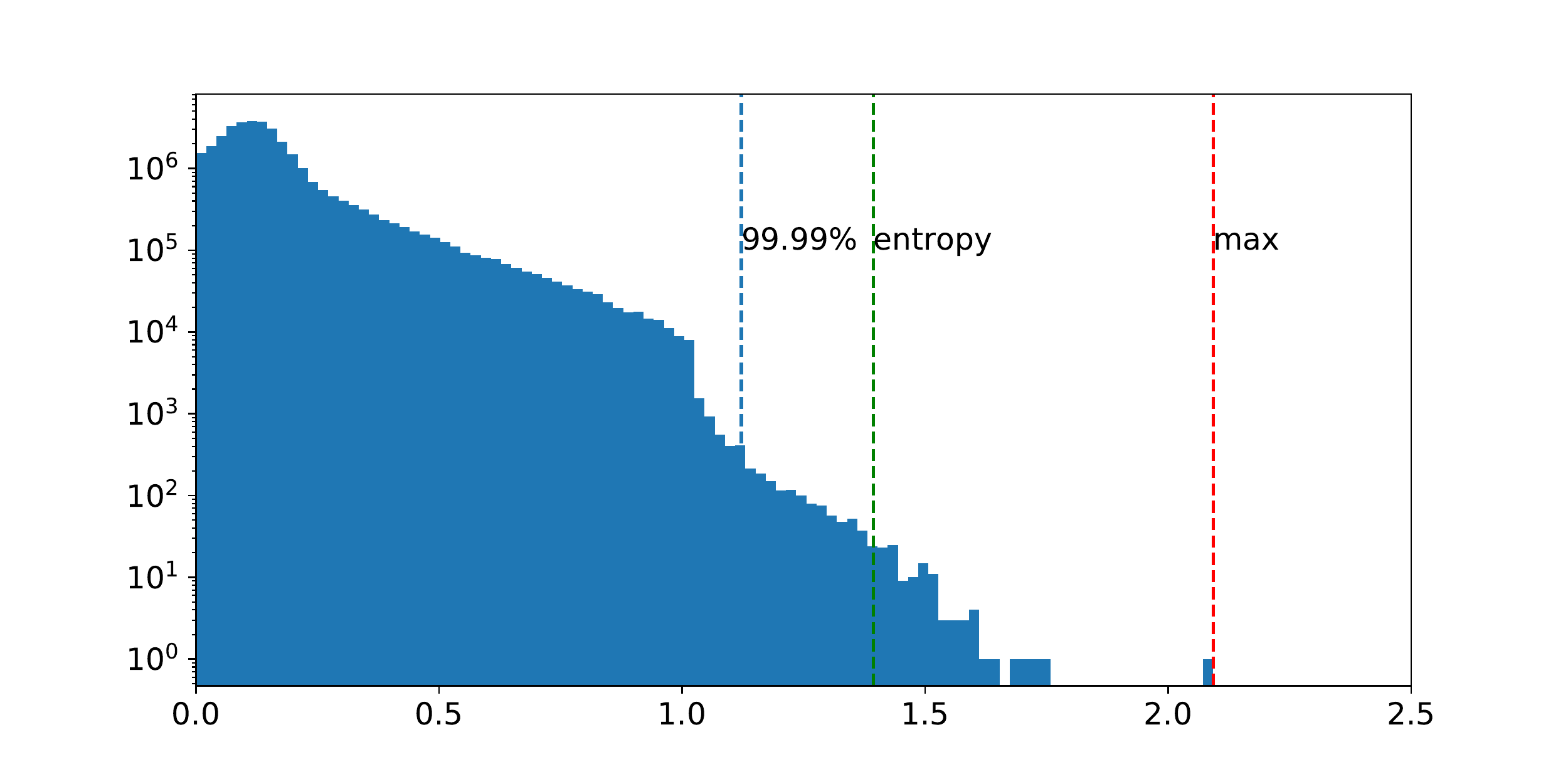}
\caption{Histogram of input activations to layer 3 in ResNet50  and calibrated ranges}
\label{fig:calib}
\end{figure}

Calibration is the process of choosing $\alpha$ and $\beta$ for model weights and activations.
For simplicity we describe calibration of a symmetric range, as needed for scale quantization. 
In this paper we consider three calibration methods: 

\texttt{Max}: Use the maximum absolute value seen during calibration~\cite{vanhoucke2011improving}.

\texttt{Entropy}: Use KL divergence to minimize information loss between
the original floating-point values and values that could be represented by the quantized format. 
This is the default method used by TensorRT~\cite{TrtQuantGtc}.

\texttt{Percentile}: Set the range to a percentile of the distribution of absolute values seen during calibration~\cite{mckinstry2018discovering}. For example, 99\% calibration would clip 1\% of the largest magnitude values.

Figure~\ref{fig:calib} shows a log scaled histogram of activations feeding into layer1.0.conv2 of ResNet50. 
Ranges derived from max, entropy, and 99.99\% percentile calibration are shown with dashed lines. 
Note that the activations are strictly non-negative because this layer directly follows a ReLU activation function~\cite{nair2010rectified}.
Max calibration represents the largest value in the distribution, maintaining the full range while having low precision.
Clipping the distribution trades off a large clipping error on a few outlier values for smaller rounding errors on a majority of the values.
Both entropy and percentile calibration clip some outlier values in order to increase the resolution of inlier values.

\section{Post Training Quantization}
\label{sec:experiments}

In this section we evaluate various Post Training Quantization (PTQ) parameter choices, as described in Section~\ref{sec:fundamentals}.
Quantization parameters are calibrated offline by processing the trained model weights and activations generated by running inference on a sample dataset, no further training is involved. 
These quantization parameters are evaluated on a variety of neural network tasks and models, summarized in Table~\ref{table:models}. More details on these networks can be found in Appendix~\ref{sec:eval_details}. 
The selected models comprise multiple types of network architectures: convolutional feed forward networks, recurrent networks, and attention-based networks. We report accuracy metrics computed on the evaluation set of the corresponding dataset. Metrics for all tasks are reported as percentages, where higher is better and 100\% is a perfect score. 
For metric consistency, we report word accuracy (WAcc) for speech recognition instead of the more commonly used Word Error Rate (WER), where $\mathrm{WAcc} = 100\% - \mathrm{WER}$. Note that accuracy metrics for different tasks are computed in very different ways, thus it is not meaningful to compare absolute changes in accuracy when quantizing different models. Therefore, when discussing accuracy impact we will refer to the relative accuracy change, computed by $(\mathrm{acc}_{int8} - \mathrm{acc}_{fp32})/\mathrm{acc}_{fp32}$.

Our experiments are conducted using PyTorch~\cite{paszke2019pytorch}, with custom quantization operations. We focus on quantizing the computationally intensive operations, including  convolutions, linear (fully-connected) layers, LSTM cells, projection layers and other matrix multiplications. Most of the other layers, such as softmax and batch normalization, are not quantized unless stated otherwise. An operation is quantized by quantizing all of its inputs (e.g. weights and activations). The output of a quantizated operation is not quantized to int8 because the operation that follows it may require higher precision, e.g. nonlinear operations.
Furthermore, consecutive operations can be executed with a fused implementation, avoiding memory reads and writes for the intermediate values.
Therefore we leave quantization of the output activations to the input of the next operation.
Appendix~\ref{sec:bn_fold} discusses how batch normalization can be eliminated by folding it into the preceding layer for inference.

\begin{table}
\centering
\begin{tabular}{lllll}

\toprule
Task & Model                       & Accuracy  & Metric    & Dataset (evaluation set)     \\ 
\midrule
\midrule
\multirow{10}{*}{Classification} &  MobileNet v1  &     71.88 &  \multirow{10}{*}{Top\-1} & \multirow{10}{*}{ImageNet 2012 (val)} \\
& MobileNet v2  &     71.88 &                         &                                           \\
& ResNet50 v1.5 &     76.16 &                         &                                           \\
& ResNet152 v1.5 &     78.32 &                         &                                           \\
& Inception v3 &     77.34 &                         &                                           \\
& Inception v4 &     79.71 &                         &                                           \\
& ResNeXt50 &     77.61 &                         &                                           \\
& ResNeXt101 &     79.30 &                         &                                           \\
& EfficientNet b0 &     76.85 &                         &                                           \\
& EfficientNet b3 &     81.61 &                         &                                           \\
\midrule
\multirow{3}{*}{Detection} & Faster R-CNN & 36.95 & \multirow{3}{*}{mAP} &  \multirow{3}{*}{COCO 2017 (val)} \\
&        Mask R-CNN  &     37.89 &                         &                                           \\
&  Retinanet  &     39.30 &                         &                                           \\
\midrule
\multirow{2}{*}{Segmentation} & FCN &     63.70 &    \multirow{2}{*}{mIoU} &                      \multirow{2}{*}{COCO 2017 (val)} \\
&             DeepLabV3 &     67.40 &                         &                                           \\
\midrule
\multirow{2}{*}{Translation} &  GNMT &     24.27 &    \multirow{2}{*}{BLEU} &  \multirow{2}{*}{WMT16 en-de (newtest2014)} \\
&   Transformer &     28.27 &                         &                                           \\
\midrule
Speech Recognition &  Jasper &     96.09 (3.91) &                     WAcc (WER) &                    LibriSpeech (test-clean) \\
\midrule
Language model &     BERT Large &     91.01 &  F1 &                                Squad v1.1 (dev) \\
\bottomrule
\end{tabular}
\caption{Summary of networks and pre-trained model accuracy}
\label{table:models}
\end{table}

\subsection{Weight Quantization}
\label{sec:wt_only_quant}

\begin{table}
\centering
\begin{tabular}{l|c|cccc}
\toprule
Model &   fp32 &  Per-channel &  Per-channel fold BN &  Per-tensor &  Per-tensor fold BN \\
\midrule
MobileNet v1 &  71.88 &        71.59 &                71.59 &       69.58 &               66.88 \\
MobileNet v2 &  71.88 &        71.61 &                71.61 &       71.12 &               70.21 \\
ResNet50 v1.5 &  76.16 &        76.14 &                76.14 &       75.83 &               75.84 \\
ResNeXt50 &  77.61 &        77.62 &                77.62 &       77.48 &               77.45 \\
EfficientNet b0 &  76.85 &        76.72 &                76.72 &       76.68 &               12.93 \\
\bottomrule
\end{tabular}
\caption{Accuracy with int8 quantization of weights only: per-tensor vs per-channel granularity. Fold BN indicates batch norms were folded into the preceding convolution before quantization}
\label{table:wt_only_fold_bn}
\end{table}

\begin{table}
\centering
\begin{tabular}{l|r|rr||l|r|rr}
\toprule
           Model &   fp32 &  Accuracy &  Relative &          Model  &  fp32  &  Accuracy & Relative \\
\midrule
    MobileNet v1 &  71.88 &     71.59 &  -0.40\% &  Faster R-CNN &   36.95 &       36.86 &   -0.24\% \\
    MobileNet v2 &  71.88 &     71.61 &  -0.38\% &    Mask R-CNN &   37.89 &       37.84 &   -0.13\% \\
   ResNet50 v1.5 &  76.16 &     76.14 &  -0.03\% &     Retinanet &   39.30 &       39.20 &   -0.25\% \\
  ResNet152 v1.5 &  78.32 &     78.28 &  -0.05\% &           FCN &   63.70 &       63.70 &    0.00\% \\
    Inception v3 &  77.34 &     77.44 &   0.13\% &     DeepLabV3 &   67.40 &       67.40 &    0.00\% \\
    Inception v4 &  79.71 &     79.64 &  -0.09\% &          GNMT &   24.27 &       24.41 &    0.58\% \\
       ResNeXt50 &  77.61 &     77.62 &   0.01\% &   Transformer &   28.27 &       28.58 &    1.10\% \\
      ResNeXt101 &  79.30 &     79.29 &  -0.01\% &        Jasper &   96.09 &       96.10 &    0.01\% \\
 EfficientNet b0 &  76.85 &     76.72 &  -0.17\% &    Bert Large &   91.01 &       90.94 &   -0.08\% \\
 EfficientNet b3 &  81.61 &     81.55 &  -0.07\% &               &         &             &           \\
\bottomrule 
\end{tabular}
\caption{Accuracy with int8 quantization of weights only: per-channel granularity, max calibration}
\label{table:wt_only_max_calib}
\end{table}

We first evaluate weight quantization in isolation, since their values do not depend on network inputs, and demonstrate that max calibration is sufficient to maintain accuracy for int8 weights.
Table~\ref{table:wt_only_fold_bn} compares the accuracy impact of the per-tensor and per-channel quantization granularities, which in Section~\ref{sec:granularity} were shown to require minimal compute overheads. While per-tensor quantization results in substantial accuracy losses for some networks, accuracy loss is more pronounced and even catastrophic for EfficientNet once batch-normalization (BN) parameters are folded into convolution layers. BN folding (Appendix~\ref{sec:bn_fold}) is a common technique to speed up inference as it completely eliminates this memory-limited operation without changing the underlying mathematics. However, as BN parameters are learned per channel, their folding can result in significantly different weight value distributions across channels. Fortunately, as Table~\ref{table:wt_only_fold_bn} shows, per-channel quantization granularity maintains model accuracy even with BN folding. 
Table~\ref{table:wt_only_max_calib} reports per-channel (per-column for linear layers) granularity and indicates that max calibration is sufficient to maintain accuracy when quantizing weights to int8. The rest of the experiments in this paper use per-channel max calibration for weights.

\subsection{Activation Quantization}
\label{sec:full_quant}

Table~\ref{table:post_quant} shows activation quantization results for different calibration methods: max, entropy and percentiles from 99.9\% to 99.9999\%.
Details on activation calibration can be found in Appendix~\ref{sec:eval_details}.
In all cases, weights were quantized per-channel with max calibration as described in Section~\ref{sec:wt_only_quant}.

For most of the networks, there is at least one activation calibration method that achieves acceptable accuracy, except for MobileNets, EfficientNets, Transformer and BERT where the accuracy drop is larger than 1\%.
Max calibration leads to inconsistent quality across various networks, leading to particularly large accuracy drops for Inception v4, EfficientNets and Transformer, presumably due to their outlier values. 
99.9\% percentile calibration clips the large magnitude values too aggressive and leads to significant accuracy drops on most networks.
The best post training quantization results are achieved with
entropy, 99.99\%, or 99.999\% percentile calibrations, though no single calibration is best for all networks.

\begin{table}
  \centering
  \begin{tabular}{l|c|cccccc}
    \toprule
Models                         & fp32  & Max   & Entropy & 99.9\% & 99.99\% & 99.999\% & 99.9999\% \\ \midrule
MobileNet v1 & 71.88 & 69.51 & 70.19   & \textbf{70.39}  & 70.29   & 69.97    & 69.57     \\
MobileNet v2 & 71.88 & 69.41 & 70.28   & 70.68  & \textbf{71.14}   & 70.72    & 70.23     \\
ResNet50 v1.5 & 76.16 & 75.82 & \textbf{76.05}   & 75.68  & 75.98   & 75.97    & 76.00     \\
ResNet152 v1.5 & 78.32 & 77.93 & \textbf{78.21}   & 77.62  & 78.17   & 78.17    & 78.19     \\
Inception v3 & 77.34 & 72.53 & \textbf{77.54}   & 76.21  & 77.52   & 77.43    & 77.37     \\
Inception v4 & 79.71 & 0.12  & 79.60   & 78.16  & \textbf{79.63}   & 79.12    & 71.19     \\
ResNeXt50  & 77.61 & 77.31 & \textbf{77.46}   & 77.04  & 77.39   & 77.45    & 77.39     \\
ResNeXt101 & 79.30 & 78.74 & 79.09   & 78.77  & 79.15   & \textbf{79.17}    & 79.05     \\
EfficientNet b0 & 76.85 & 22.3  & \textbf{72.06}   & 70.87  & 68.33   & 51.88    & 42.49     \\
EfficientNet b3 & 81.61 & 54.27 & 76.96   & 77.80  & \textbf{80.28}   & 80.06    & 77.13     \\
Faster R-CNN & 36.95 & 36.38 & \textbf{36.82}   & 35.22  & 36.69   & 36.76    & 36.78     \\
Mask R-CNN  & 37.89 & 37.51 & 37.75   & 36.17  & 37.55   & 37.72    & \textbf{37.80}     \\
Retinanet & 39.30 & 38.90 & 38.97   & 35.34  & 38.55   & \textbf{39.19} & \textbf{39.19}     \\
FCN & 63.70 & 63.40 & 64.00   & 62.20  & 64.00   & \textbf{63.90}    & 63.60     \\
DeepLabV3 & 67.40 & 67.20 & 67.40   & 66.40  & 67.40   & \textbf{67.50}    & 67.40     \\
GNMT & 24.27 & 24.31 & \textbf{24.53}   & 24.34  & 24.36   & 24.38    & 24.33     \\
Transformer & 28.27 & 21.23 & 21.88   & 24.49  & \textbf{27.71}   & 20.22    & 20.44     \\
Jasper & 96.09 & 95.99 & ~\textbf{96.11}   & 95.77  & 96.09   & 96.09    & 96.03      \\
BERT Large      & 91.01 & 85.92 & 37.40    & 26.18  & 89.59   & \textbf{90.20} & 90.10      \\
    \bottomrule
  \end{tabular}
 \caption{Post training quantization accuracy. Weights use per-channel or per-column max calibration. Activations use the calibration listed. Best quantized accuracy per network is in bold.}
  \label{table:post_quant}
\end{table}

\section{Techniques to Recover Accuracy}
\label{sec:acc-recovery}
While many networks maintain accuracy after post training quantization, there are cases where accuracy loss is substantial.
A number of techniques are available to recover accuracy. The simplest one is partial quantization, described in Section~\ref{sec:sensitivity}, which leaves the most sensitive layers unquantized. 
One also has an option to train networks with quantization, as described in Section~\ref{sec:qat}. Finally, there are also approaches that jointly learn the model weights and quantization parameters.

\subsection{Partial Quantization}
\label{sec:sensitivity}

Often just a few quantized layers contribute to most of the accuracy loss of a quantized model.
We can trade off some performance to increase accuracy by leaving these sensitive layers unquantized (i.e. leaving their inputs and computation in floating-point).
Since quantization of one layer affects the inputs of others, finding the optimal set of layers to quantize can require evaluating an exponential number of configurations. Instead, we propose using a one-at-a-time sensitivity analysis as a more tractable approach to infer which layers contribute most to the accuracy drop.

During sensitivity analysis a single layer is quantized at a time, and model accuracy is evaluated.
We refer to layers that result in lower accuracy when quantized as being more ``sensitive'' to quantization.
We sort the layers in descending order of sensitivity, and skip quantization of the most sensitive layers until the desired accuracy is achieved.

\begin{figure}[t]
\centering
\includegraphics[width=0.75\linewidth, trim=0 0 0 0, clip]{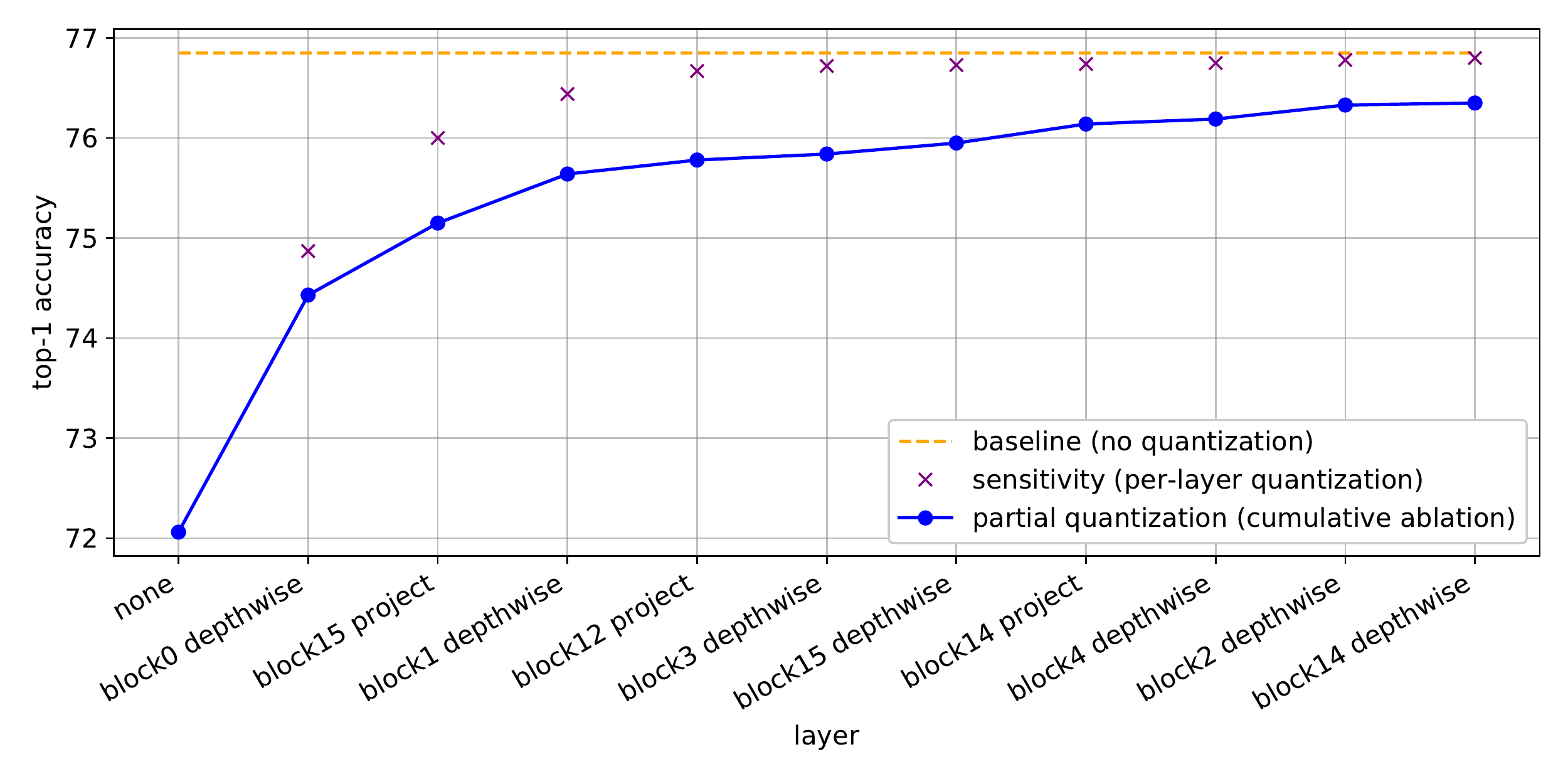}
\caption{Partial quantization of EfficientNet b0, showing the 10 most sensitive layers in order of increasing accuracy.
Sensitivity shows the accuracy from the sensitivity analysis when only the corresponding layer inputs are quantized.
Partial quantization shows the accuracy when the corresponding layer, and all layers to the left, are not quantized.
}
\label{fig:enb0_sensitivty}
\end{figure}

Figure~\ref{fig:enb0_sensitivty} shows an example of sensitivity analysis and partial quantization of EfficientNet b0. 
Starting from entropy calibration, we quantize one layer at a time and evaluate accuracy.
For clarity we are only showing the 10 most sensitive layers.
In this example, skipping the 10 most sensitive layers reduces the relative top-1 accuracy drop to 0.65\%.
Since there are 82 convolution layers, keeping 10 in floating-point while quantizing the remaining 72 maintains most of the performance benefit.

\begin{table}[t]
  \centering
  \begin{tabular}{l|c|c|cc|cc}
\toprule
& fp32 & & \multicolumn{2}{c|}{Full int8}  & \multicolumn{2}{c}{Partial int8} \\
Model &  Accuracy &  Calibration &  Total quantized layers &   Accuracy &  Skipped layers &   Accuracy \\
\midrule

MobileNet v1 &     71.88 &          max &                                28 &      69.51 &              2 &      71.50 \\
EfficientNet b0 &     76.85 &      entropy &                                82 &      72.06 &             10 &      76.35 \\
EfficientNet b3 &     81.61 &      99.99\% &                               131 &      76.96 &              3 &      81.27 \\
Transformer &     28.27 &          max &                               121 &      21.23 &              5 &      28.20 \\
BERT large &     91.01 &          max &                               244 &      85.92 &            141 &      90.41 \\
\bottomrule
\end{tabular}
\caption{Partial post training quantization}
\label{table:sensitivity}
\end{table}

As reported in Table~\ref{table:post_quant}, MobileNet v1, EfficientNets, Transformer, and BERT all incurred a substantial loss in accuracy when quantized with various calibrations. We list the results of partial quantization for these networks in Table~\ref{table:sensitivity}. With the exception of BERT, these networks need to skip quantization of only a few of the most-sensitive layers to recover accuracy to within 1\% of the fp32 accuracy. For BERT, sensitivity analysis does not reveal any particular layer that contributes more to the accuracy drop.
As a result we cannot identify a small subset of layers to leave in floating-point. 
To address this we need to consider different approaches. 
Section~\ref{sec:qat}, incorporates quantization with training to recover accuracy.
Additionally, Appendix~\ref{sec:asymmetric} examines the GELU activation function in BERT and presents a simple augmentation to significantly improve post training quantization accuracy.

\subsection{Quantization-Aware Training}
\label{sec:qat}

\begin{figure}[t]
\centering
\includegraphics[width=0.4\linewidth, trim=0 0 0 0, clip]{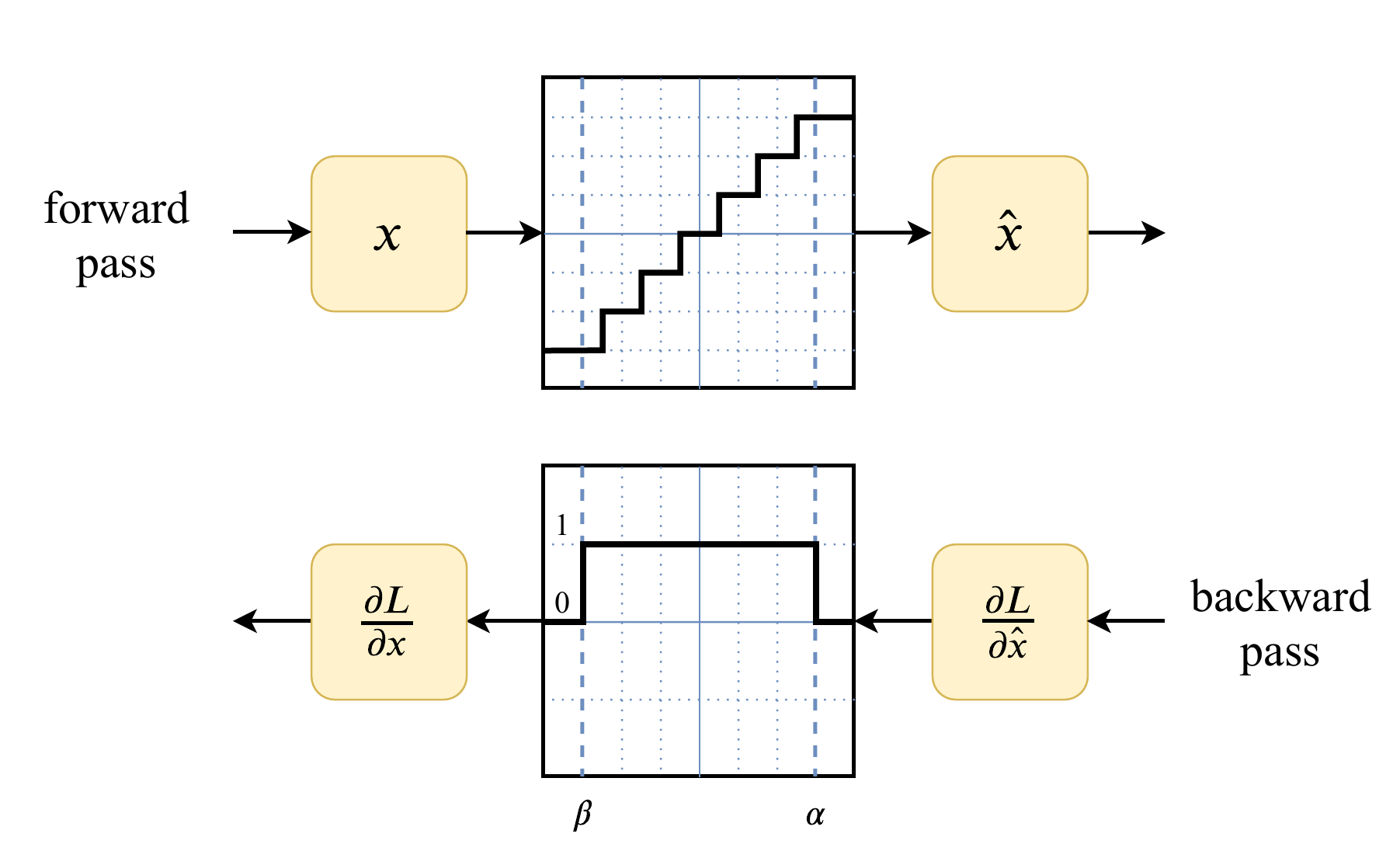}
\caption{3-bit fake quantization forward and backward pass with STE derivative approximation.}
\label{fig:ste}
\end{figure}

Quantization Aware Training (QAT) describes the technique of inserting quantization operations in to the neural network before training or fine-tuning, to allow the network to adapt to the quantized weights and activations. 
Appendix~\ref{sec:intuition} illustrates how this can lead to a better result. 
We apply QAT to fine-tuning as it has been shown that starting from a pre-trained network and fine-tuning leads to better accuracy~\cite{mishra2017apprentice,krishnamoorthi2018quantizing} and requires significantly fewer iterations~\cite{mckinstry2018discovering}.
This also allows us to leverage the calibrated pre-trained models from Section~\ref{sec:experiments}. 
Note that we keep the quantization ranges fixed throughout fine-tuning.
Another approach is to learn the ranges, which we evaluate in Section~\ref{sec:pact}.

A common approach to implementing QAT is to insert fake quantization, also called simulated quantization~\cite{krishnamoorthi2018quantizing}, operations into a floating-point network.
Equation ~\ref{eq:fake} defines fake quantization as a quantize and dequantize operation that produces an approximate version of the input, $\hat{x} \approx x$, where $x$ and $\hat{x}$ are both floating-point values.

\begin{equation}
\hat{x} = \dequantize(\quantize(x, b, s), b, s) 
\label{eq:fake}
\end{equation}

We add fake quantization operations at the inputs of the operation we wish to quantize to simulate the effects of quantization.
Recall the matrix multiplication example in Section~\ref{sec:granularity}.
Equation~\ref{eq:dot-product-granularity} is effectively a fake quantized matrix multiplication.
After training, we transform the network to enable a quantized integer matrix multiply as shown in Equation~\ref{eq:scale_only_quant}.

One challenge to training in floating-point with quantization is that the quantization operation's derivative is undefined at the step boundaries and zero everywhere else. The derivative is required to compute loss gradients on the backward pass of each training iteration.
QAT addresses this by using the 
Straight-through Estimator (STE)~\cite{bengio2013estimating} as shown in Figure~\ref{fig:ste}. 
As defined in Equation~\ref{eq:ste}, STE approximates the derivative of the fake quantization function to be 1 for inputs in the representable range $[\beta, \alpha]$ as defined in Section~\ref{sec:range_map}.

\begin{equation}
 \frac{d\hat{x}}{dx} =
\begin{cases}
    0, & y < \beta \\
    1, &  \beta \le y \le \alpha \\
    0, & y > \alpha
\end{cases}
\label{eq:ste}
\end{equation}

Table~\ref{table:best_quant} summarizes the best results of both post training quantization and fine-tuned quantization. 
PTQ best reports the best result for each quantized network in Table~\ref{table:post_quant} and the corresponding calibration.
QAT reports the accuracy after fine-tuning using the best calibration as determined by PTQ.
Details of the finetuning methodology and the complete set of QAT results can be found in Appendix~\ref{sec:quantized-training}.

As Table~\ref{table:best_quant} shows, quantization-aware fine-tuning improves accuracy in most cases, the only exceptions being ResNeXt-101, Mask R-CNN, and GNMT where post training quantization achieves a marginally better result. 
It is worth noting that for all 3 of these cases the differences in accuracy are essentially at the noise level (differences in accuracy one would observe when training from different random initializations). 
We do not interpret these cases as evidence that fine-tuning reduces accuracy, they are more likely to indicate that fine-tuning does not appreciably change accuracy beyond run-to-run variation. 
Likewise, we do not interpret cases where accuracy is higher than fp32 as quantization acting as a regularizer, it is more likely to be noise or the result of the additional fine-tuning. 
EfficientNet b3 is another case worth examining - as our code did not have auto augmentation~\cite{cubuk2019autoaugment}, used to train the original model, fine-tuning even in fp32 causes a slight accuracy drop to 81.3. Nevertheless, with fine-tuning all networks were able to maintain their accuracy well within 1\% of the original pre-trained fp32 model.

\begin{table}[t]
  \centering
  \begin{tabular}{l|c|lcr|cr}
\toprule
                 &      fp32 & \multicolumn{3}{c|}{PTQ best}  & \multicolumn{2}{c}{QAT} \\
           Model &  Accuracy &                  Calibration &   Accuracy &   Relative &          Accuracy &   Relative \\
\midrule
    MobileNet v1 &     71.88 &                       99.9\% &      70.39 &    -2.07\% &             72.07 &     0.26\% \\
    MobileNet v2 &     71.88 &                      99.99\% &      71.14 &    -1.03\% &             71.56 &    -0.45\% \\
   ResNet50 v1.5 &     76.16 &                      Entropy &      76.05 &    -0.14\% &             76.85 &     0.91\% \\
  ResNet152 v1.5 &     78.32 &                      Entropy &      78.21 &    -0.14\% &             78.61 &     0.37\% \\
    Inception v3 &     77.34 &                      Entropy &      77.54 &     0.26\% &             78.43 &     1.41\% \\
    Inception v4 &     79.71 &                      99.99\% &      79.63 &    -0.10\% &             80.14 &     0.54\% \\
       ResNeXt50 &     77.61 &                      Entropy &      77.46 &    -0.19\% &             77.67 &     0.08\% \\
      ResNeXt101 &     79.30 &                     99.999\% &      79.17 &    -0.16\% &             79.01 &    -0.37\% \\
 EfficientNet b0 &     76.85 &                      Entropy &      72.06 &    -6.23\% &             76.95 &     0.13\% \\
 EfficientNet b3 &     81.61 &                      99.99\% &      80.28 &    -1.63\% &             81.07 &    -0.66\% \\
    Faster R-CNN &     36.95 &                      Entropy &      36.82 &    -0.35\% &             36.76 &    -0.51\% \\
      Mask R-CNN &     37.89 &                    99.9999\% &      37.80 &    -0.24\% &             37.75 &    -0.37\% \\
       Retinanet &     39.30 &                     99.999\% &      39.19 &    -0.28\% &             39.25 &    -0.13\% \\
             FCN &     63.70 &                      Entropy &      64.00 &     0.47\% &             64.10 &     0.63\% \\
       DeepLabV3 &     67.40 &                     99.999\% &      67.50 &     0.15\% &             67.50 &     0.15\% \\
            GNMT &     24.27 &                      Entropy &      24.53 &     1.07\% &             24.38 &     0.45\% \\
     Transformer &     28.27 &                      99.99\% &      27.71 &    -1.98\% &             28.21 &    -0.21\% \\
          Jasper &     96.09 &                      Entropy &      96.11 &     0.02\% &             96.10 &     0.01\% \\
      BERT Large &     91.01 &                     99.999\% &      90.20 &    -0.89\% &             90.67 &    -0.37\% \\
\bottomrule

\end{tabular}
\caption{Summary of Post Training Quantization and Quantization Aware Training. PTQ best reports the best accuracy and corresponding calibration for each model. QAT reports accuracy after fine-tuning starting from the best PTQ model.}
\label{table:best_quant}
\end{table}

\subsection{Learning Quantization Parameters}
\label{sec:pact}

\begin{table}[tb]
\centering
\begin{tabular}{l|c|cc|cc}
\toprule
Models       &   fp32 &  Fixed max &  Learned from max &  Fixed best &  Learned from best \\
\midrule
Inception v3 &  77.34 &    76.43 &     78.33 &     78.43 &      78.50 \\
Inception v4 &  79.71 &    68.38 &     73.88 &     80.14 &      80.00 \\
Faster R-CNN &  36.95 &    36.62 &     36.68 &     36.76 &      36.81 \\
FCN          &  63.70 &    63.40 &     63.50 &     64.10 &      64.00 \\
Transformer  &  28.27 &    28.42 &     28.08 &     28.21 &      28.39 \\
Jasper       &  96.09 &    96.11 &     96.05 &     96.10 &      96.06 \\
BERT Large   &  91.01 &    90.29 &     90.55 &     90.67 &      90.61 \\
\bottomrule
\end{tabular}
\caption{Learned and fixed range fine-tuning accuracy. Activation ranges initialized to max and best PTQ accuracy}
\label{table:pact}
\end{table}

While the techniques described in the previous sections relied on quantization parameters calibrated on the pre-trained network, it is also possible to jointly learn the quantization parameters along with the model weights. PACT~\cite{choi2018pact} proposed learning the ranges for activation quantization during training. 
In this section we adopt PACT as an enhancement to our quantization aware fine-tuning procedure.
We follow the same fine-tuning schedule as before, described in Appendix~\ref{sec:eval_details}, but allow the ranges of each quantized activation tensor to be learned along with the weights, as opposed to keeping them fixed throughout fine-tuning.

Table~\ref{table:pact} shows a selection of networks fine-tuned with fixed and learned activation ranges for different initial calibrations. 
The ``best'' calibration refers to the calibration that produced the best accuracy with PTQ, as shown in Table~\ref{table:post_quant}.
When the activation quantization is initialized with max calibration, learning the range results in higher accuracy than keeping it fixed for most networks. In particular it results in substantial accuracy improvements where fixed max ranges resulted in a significant accuracy drop.
However, when activation ranges are initialized the to the best calibration for each network, learning the ranges yield very similar results to fixed ranges.
This suggests that learning the ranges does not offer additional benefit for int8 over QAT if activation ranges are already carefully calibrated.
However, this may not be the optimal application of PACT. 
Comparing the learned range results on Inception v4 suggest that when starting from max, the network was not able to learn a good activation ranges in the given fine-tuning schedule.
We expect that PACT would be able to learn a better range with longer fine-tuning, or a separate optimization schedule and hyperparameters for the range parameters, such and learning rate and weight decay.

\section{Recommended Workflow}
\label{sec:workflow}

Based on the results in Sections~\ref{sec:experiments} and~\ref{sec:acc-recovery}, we recommend the following for int8 quantization:

\begin{itemize}
  \item Weights:
    \begin{itemize}
      \item Use scale quantization with per-column/per-channel granularity
      \item Use a symmetric integer range for quantization {[}-127, 127{]}) and max calibration
    \end{itemize}
  \item Activations:
    \begin{itemize}
      \item Use scale quantization with with per-tensor granularity
    \end{itemize}
\end{itemize}

We recommend the following procedure to quantize a pre-trained neural network.

\begin{figure}[t]
\centering
\includegraphics[scale=0.7]{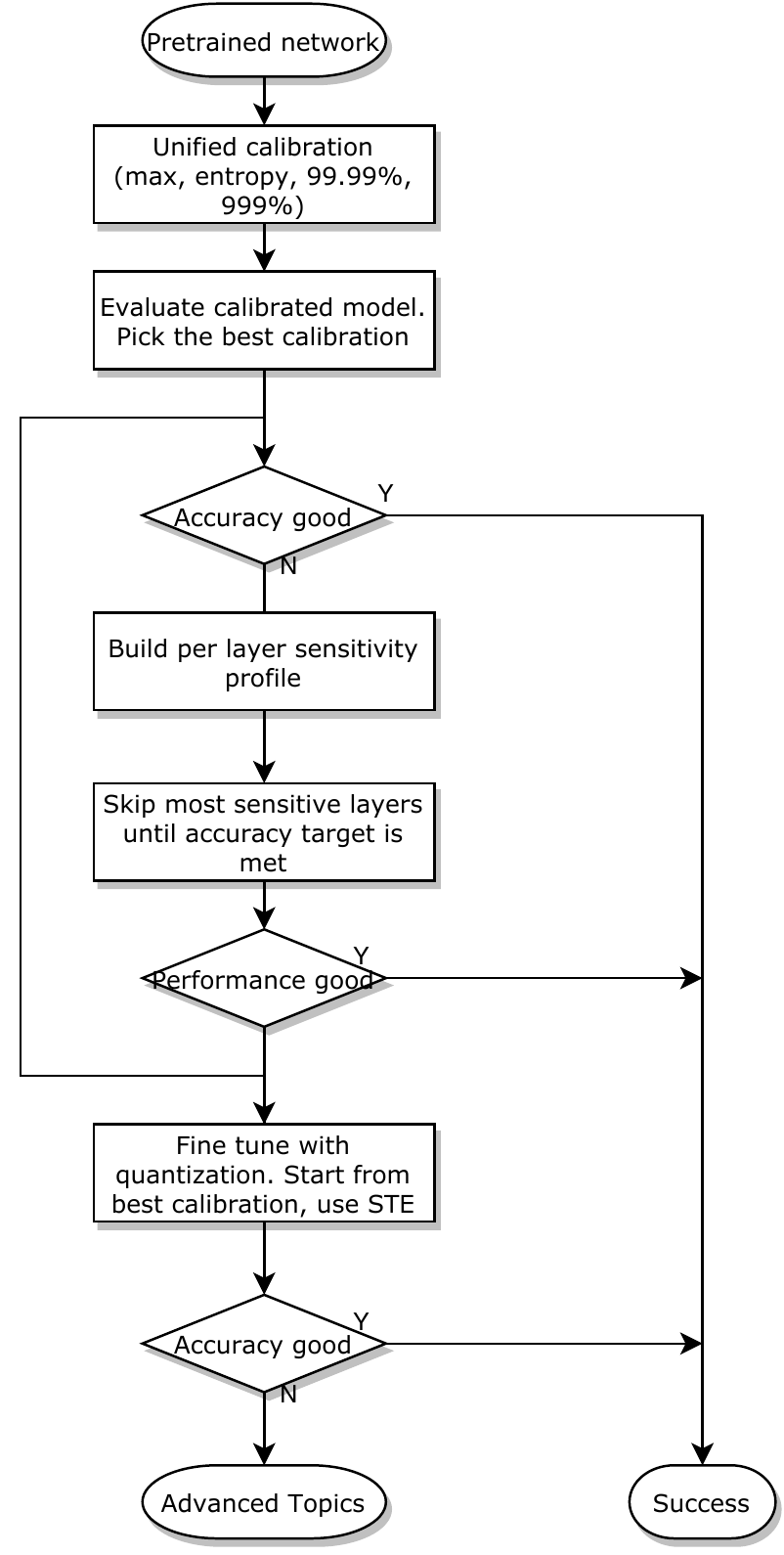}
\caption{Flow chart of our recommended quantization workflow}
\label{fig:recipe}
\end{figure}

\begin{itemize}
\item
  \texttt{PTQ}: Quantize all the computationally intensive layers (convolution, linear, matrix multiplication, etc.) and run activation calibration including max, entropy and 99.99\%, 99.999\% percentile.
  If none of the calibrations yield the desired accuracy continue to partial quantization or QAT.
\item
  \texttt{Partial Quantization}: Perform sensitivity analysis to identify the most sensitive layers and leave them in floating-point. If the impact on computational performance is not acceptable or an acceptable accuracy cannot be reached, continue to QAT.

\item
  \texttt{QAT}: Start from the best calibrated quantized model. Use QAT to fine-tune for around 10\% of the original training schedule with an annealing learning rate schedule starting at 1\% of the initial training learning rate. Refer to Appendix~\ref{sec:quantized-training} for specific hyperparameter choices. 
\end{itemize}
Figure~\ref{fig:recipe} summarizes the above workflow in a flowchart.

\section{Conclusions}

This paper reviewed the mathematical background for integer quantization of neural networks, as well as some performance-related reasons for choosing quantization parameters. We empirically evaluated various choices for int8 quantization of a variety of models, leading to a quantization workflow proposal. 
Following this workflow we demonstrated that all models we studied can be quantized to int8 with accuracy that either matches or is within 1\% of the floating-point model accuracy. This included networks that are challenging for quantization, such as MobileNets and BERT. The workflow involves only post-training quantization, partial quantization, and quantization-aware fine-tuning techniques. Some more complex techniques, such as ADMM and distillation, were not required for int8 quantization of these models. However, these techniques should be evaluated when quantizing to even lower-bit integer representations, which we leave to future work.

\bibliographystyle{plain}

\bibliography{references}

\newpage
\appendix
\appendixpage
\addappheadtotoc

\section{Evaluation Details} \label{sec:eval_details}

\subsection{Model Definitions}

\begin{table}[bh]
\begin{tabular}{llrl}
\toprule
Model & Configuration &  Calibration samples & Source \\
\midrule
\midrule
MobileNet v1 & width\_mult=1.0 & 1024 & github.com/marvis/pytorch-mobilenet                                      \\
\midrule
MobileNet v2 & width\_mult=1.0 & 1024 & \multirow{7}{*}{github.com/pytorch/vision} \\
ResNet50 v1.5 &                       &               1024 &                                              \\
ResNet152 v1.5 &                       &               1024 &                                              \\
Inception v3 &  &               1024 &                                              \\
Inception v4 &  &               1024 &                                              \\
ResNeXt50 &                 32x4d &               1024 &                                              \\
ResNeXt101 &                 32x8d &               1024 &                                              \\
\midrule
EfficientNet b0 &                       &               1024 &     \multirow{2}{*}{github.com/lukemelas/efficientnet-pytorch} \\
EfficientNet b3 & & 1024 & \\
\midrule
Faster R-CNN  &  resnet50-fpn &                512 &                    \multirow{4}{*}{github.com/pytorch/vision} \\
Mask R-CNN  &    resnet50-fpn &                512 &                                              \\
FCN &            resnet101 &                512 &                                              \\
DeepLabV3 &      resnet101 &                512 &                                              \\
\midrule
Retinanet  &     resnext101-32x4d-fpn &                512 &            github.com/open-mmlab/mmdetection \\
\midrule
Transformer & vaswani\_en\_de\_big & 14336 tokens & github.com/pytorch/fairseq \\
\midrule
GNMT &              4 layers & 128 &       \multirow{2}{*}{github.com/nvidia/deeplearningexamples} \\
Jasper &                       & full dev-clean & \\
\midrule
Bert Large & fine-tuned for QA &  128 &  github.com/huggingface/pytorch-transformers \\
\bottomrule
\end{tabular}
\caption{Network details}
\label{table:models_detail}
\end{table} 

Table~\ref{table:models_detail} lists additional details of the models listed in Table~\ref{table:models}.
We evaluated a large variety of CNNs for image classification.
MobileNets are small networks that target inference on mobile devices~\cite{howard2017mobilenets,sandler2018mobilenetv2}. They are parameterized to scale to various channel widths and image resolutions.
In this paper we evaluate the base configurations with width multiplier 1 and resolution 224x224.
We also evaluated a number of larger CNNs~\cite{he2016deep, xie2017aggregated,szegedy2016rethinking,szegedy2017inception}, including EfficientNets~\cite{efficientnet}, which achieve state-of-the-art accuracy on ImageNet.
All CNNs use 224x224 inputs except for Inception v3 and v4, which use 299x299. 
We evaluated two detection and two segmentation networks from Torchvision, and an additional segmentation network, RetinaNet~\cite{lin2017focal}. 
We evaluated two translation models, the 4 layers GNMT model~\cite{wu2016google}
and the large configuration of Transformer~\cite{Transformer2017}.
For speech recognition we evaluated Jasper
which achieves state-of-the-art WER on public speech datasets~\cite{panayotov2015librispeech}.
For language modeling we use BERT large uncased and fine-tuned for question answering.

Models were calibrated with the number of samples listed from the training set of the respective dataset listed in Table~\ref{table:models}, except for Jasper, which was calibrated on the dev set and evaluated on the test set.
PyTorch implementations of all the models along were provided by the listed source repositories.
We used the pre-trained weights provided by each repository, except for MobileNet v1 and EfficientNets where pre-trained weights were not available.
MobileNet v1 was trained using the reference training script and hyperparameters for MobileNet v2 from Torchvision.
Pre-trained weights for EfficientNets were converted to PyTorch from weights provided by TensorFlow~\cite{abadi2016tensorflow}\footnote{https://github.com/tensorflow/tpu/tree/master/models/official/efficientnet}.

\newpage
\subsection{Quantization Aware Training}
\label{sec:quantized-training}

Table~\ref{table:finetune_schedule} shows the fine-tuning hyperparameters used in the quantization aware fine-tuning experiments. 
For networks that are trained on multiple datasets (detection/segmentation networks and BERT) we only fine-tuned on the final dataset (COCO and SQuAD).
In general, only the initial learning rate value and learning rate schedule are changed from the original training session.
We fine-tune for around 1/10th of the original training steps. The fine-tuning learning rate starts at 1/100th of the initial value used in the original training session and is decayed down to 1/100th of the initial fine-tuning learning rate.
BERT is an exception. Since it pre-trains a language model and only fine-tunes on SQuAD for 2 epochs, we instead repeat the full fine-tuning schedule for QAT.
We used a cosine annealing learning rate schedule which follows the monotonically decreasing half period of the cosine function.

\begin{table}[tb]
\centering
\begin{tabular}{llcrrl}
\toprule
Task/Model & Dataset & Optimizer & Epochs & Initial learning rate & Batch size\\
\midrule
Classification      & ImageNet 2012     & SGD       & 15     & 1.0e-3                  & 256                            \\
Detection          & COCO 2017      & SGD       & 3      & 2.5e-5                & 16                             \\
Segmentation       & COCO 2017       & SGD       & 3      & 2.5e-5                & 32                             \\
Transformer & WMT16 en-de & ADAM      & 3      & 5.0e-4                  & max tokens = 3584              \\
GNMT       & WMT16 en-de & SGD       & 1      & 2.0e-3                  & 1024, max seq. length = 50 \\
Jasper        &    LibriSpeech        & NovoGrad\cite{ginsburg2019stochastic}  & 80     & 1.5e-2                 & 256                            \\
BERT         &  SQuAD v1.1  & ADAM      & 2      & 3.0e-5                  & 12, max seq. length = 384 \\  
\bottomrule
\end{tabular}
\caption{Fine-tuning schedule and configuration}
\label{table:finetune_schedule}
\end{table}

Table~\ref{table:finetune_quant} shows fine-tuned quantization accuracy for all networks and activation range calibration settings.
Note that we always use the full per-column/per-channel range for weights (max calibration).
It shows that with fine-tuning, accuracy improves for almost all the cases, especially those that suffer large accuracy drops after PTQ, for example max calibration.
For many of the models, the best PTQ calibration is also the best calibration for QAT, indicated by results that are both bold and underlined.
Even when QAT achieves higher accuracy with a different calibration, the difference in results is marginal.
This result suggests that evaluating multiple activation calibrations during PTQ is a good heuristic to choose a calibration for QAT.

\begin{table}
\centering
\begin{tabular}{l|r|cccccc}
\toprule
Models          & fp32  & Max   & Entropy & 99.9\% & 99.99\% & 99.999\% & 99.9999\% \\\midrule
MobileNet v1    & 71.88 & 71.80 & 72.11   & \underline{72.07}  & \textbf{72.14}   & 71.89    & 71.91     \\
MobileNet v2    & 71.88 & 71.11 & 71.50   & 71.48  & \underline{\textbf{71.56}}   & 71.28    & 71.34     \\
ResNet50 v1.5   & 76.16 & 76.68 & \underline{\textbf{76.85}}   & 76.59  & 76.67   & 76.77    & 76.81     \\
ResNet152 v1.5  & 78.32 & 78.64 & \underline{78.61}   & 78.61  & \textbf{78.69}   & 78.65    & 78.65     \\
Inception v3    & 77.34 & 76.43 & \underline{78.43}   & 78.33  & \textbf{78.49}   & 78.36    & 78.38     \\
Inception v4    & 79.71 & 68.38 & 80.07   & 80.01  & \underline{\textbf{80.14}}   & 79.94    & 78.82     \\
ResNeXt50       & 77.61 & 77.38 & \underline{\textbf{77.67}}   & 77.48  & 77.56   & 77.51    & 77.51     \\
ResNeXt101      & 79.30 & 78.98 & 78.99   & 79.00  & 78.99  & \underline{79.01}    & \textbf{79.04}     \\
EfficientNet b0 & 76.85 & 76.16 & \underline{76.95}   & 76.85  & \textbf{77.09}   & 76.98    & 76.63     \\
EfficientNet b3 & 81.61 & 80.51 & 80.63   & 80.62  & \underline{81.07}   & \textbf{81.09}    & 80.92     \\
Faster R-CNN    & 36.95 & 36.62 & \underline{36.76}   & 36.31  & 36.76   & \textbf{36.83}    & 36.76     \\
Mask R-CNN      & 37.89 & 37.63 & 37.74   & 37.26  & 37.67   & \textbf{37.76}    & \underline{37.75}     \\
Retinanet       & 39.30 & 39.03 & 39.11   & 37.76  & 38.97   & \underline{\textbf{39.25}}    & \underline{39.20}     \\
FCN             & 63.70 & 63.40 & \underline{64.10}   & 63.90  & \textbf{64.20}   & 63.90    & 63.40     \\
DeepLabV3       & 67.40 & 67.10 & 67.30   & 66.90  & 67.20   & \underline{\textbf{67.50}}    & 67.20     \\
GNMT            & 24.27 & \textbf{24.49} & \underline{24.38}   & 24.35  & 24.41   & 24.48    & 24.35     \\
Transformer     & 28.27 & 28.42 & \textbf{28.46}   & 28.23  & \underline{28.21}   & 28.04    & 28.10     \\
Jasper          & 96.09 & \textbf{96.11} & \underline{96.10} &  95.23 &   95.94 &   96.01 &    96.08     \\
BERT Large      & 91.01 & 90.29 & 90.14   & 89.97  & 90.50   & \underline{\textbf{90.67}}    & 90.60     \\
\bottomrule
\end{tabular} \\
\caption{Fine-tuned quantization. Best accuracy in bold. Accuracy from best PTQ calibration per network underlined.}
\label{table:finetune_quant}
\end{table}

\section{Intuition for QAT}
\label{sec:intuition}

\begin{figure}[t]
\centering
\begin{subfigure}{0.3\textwidth}
\includegraphics[width=0.9\linewidth, trim=10 0 50 0, clip]{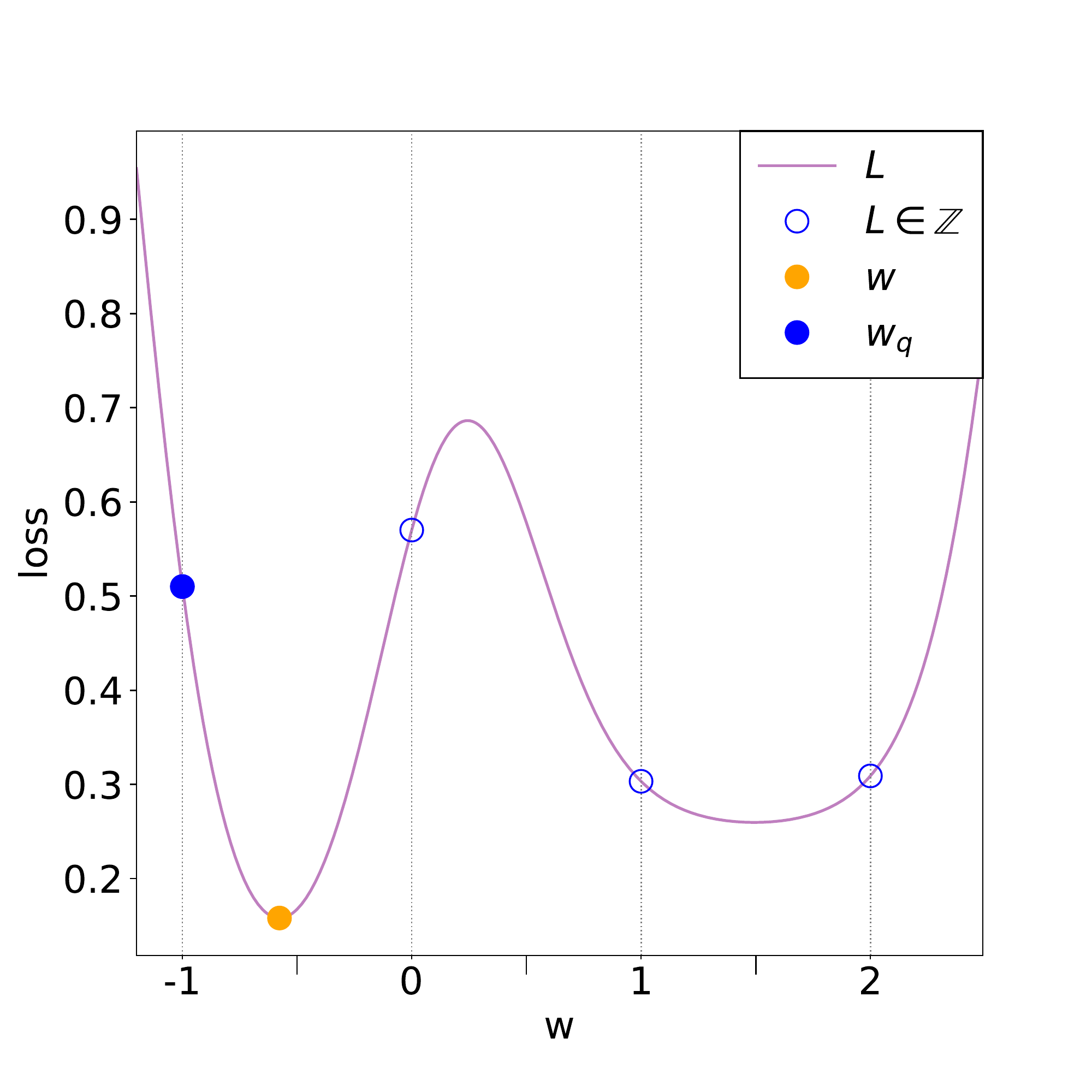}
\caption{Post training quantization}
\label{fig:qat_before}
\end{subfigure} 
\hspace{50pt}
\begin{subfigure}{0.3\textwidth}
\includegraphics[width=0.9\linewidth, trim=10 0 50 0, clip]{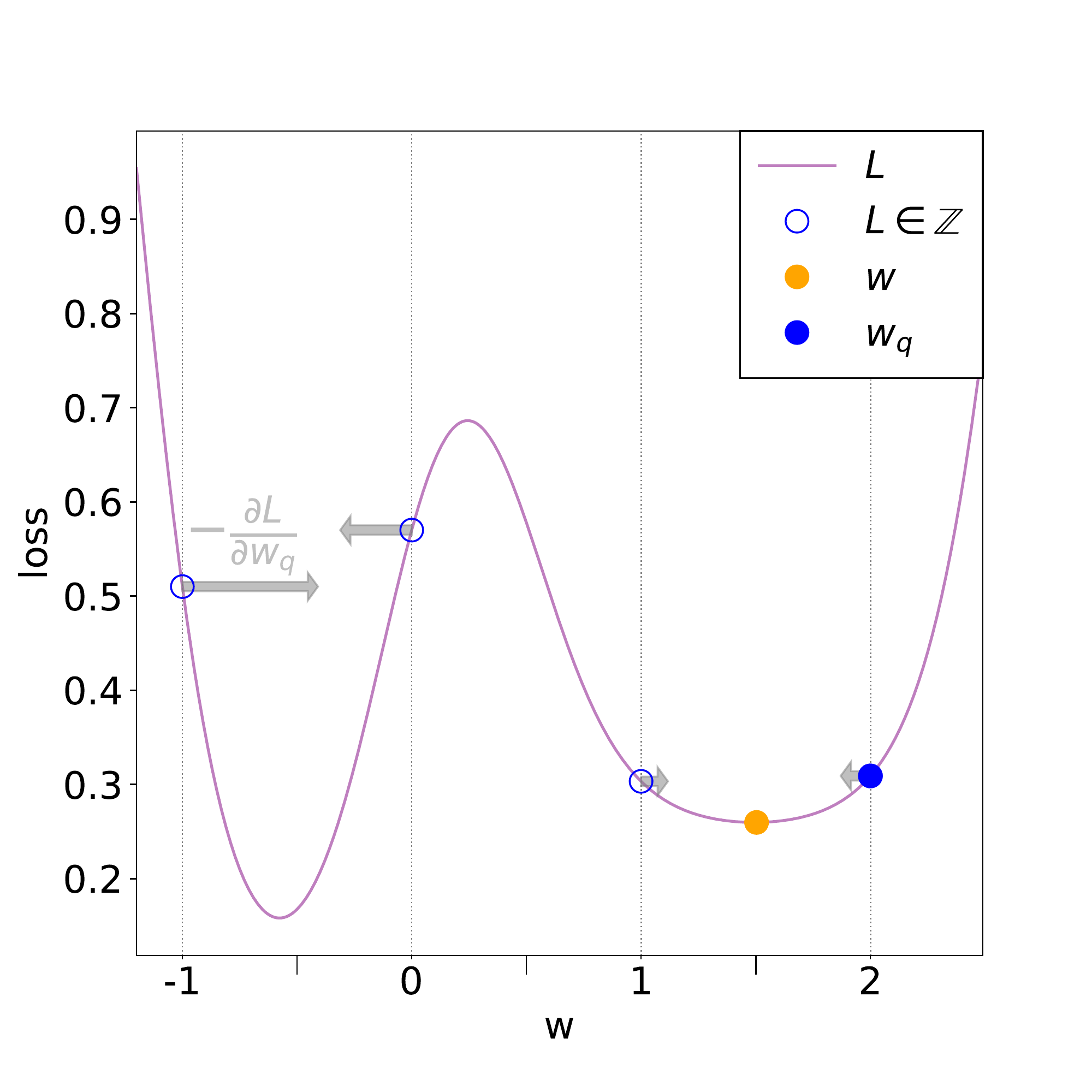}
\caption{After quantization aware fine-tuning}
\label{fig:qat_after}
\end{subfigure}
\caption{Example 1D loss function. The model, $w$, is scale quantized with scale factor 1. a) PTQ: model converges to a narrow minimum. b) QAT: model finds a wide minimum with lower loss quantization points.}
\label{fig:qat}
\end{figure}

To gain some intuition for why quantization-aware training may improve accuracy of the quantized model, consider the simple example in Figure~\ref{fig:qat}. 
Neural networks are trained by minimizing a loss function with stochastic gradient descent.
Loss gradients with respect to the network weights, $\frac{\delta L}{\delta w}$, are computed and weights are iteratively updated in the direction of the negative gradient until the model converges to some minimum. 
Figure~\ref{fig:qat_before} shows a one-dimensional loss function for a model with a single parameter, $w$, that has converged to a local minimum, $w \approx -0.5$.
When post training quantization is applied, with a scale factor of 1 for sake of example, the weight is quantized to the nearest integer, $w_q = -1$, causing a significant increase in the loss.
In such a case we say that the model has converged to a ``narrow'' minimum, since a small change in the weight leads to a large change in loss.

By training with quantization, we may potentially avoid these narrow minima by computing gradients with respect to the quantized weights, as shown in Figure~\ref{fig:qat_after}.
In doing so, narrow minima will result in larger gradients, potentially allowing the model to explore the loss landscape for ``wide''~\cite{jastrzkebski2017three} or ``flat''~\cite{hochreiter1995simplifying,lee2018retraining} minima, where quantization points have lower loss, and thus higher accuracy.

\section{Batch normalization folding} \label{sec:bn_fold}

Batch normalization folding is a common inference optimization applied to neural networks~\cite{jacob2018quantization}.
At inference, batch normalization layers performs the affine operation shown in Equation~\ref{eq:bn}:

\begin{equation}
\begin{split}
c&=\frac{\gamma}{\sqrt{Var[y]+\epsilon}} \\
d&= \beta-\frac{\gamma E[y]}{\sqrt{Var[y]+\epsilon}} \\
z&= \mathrm{BN}(y) = c \cdot y + d
\end{split}
\label{eq:bn}
\end{equation}

where $\beta$, $\gamma$, $E[y]$, and $Var[y]$ are determined during training and fixed during inference, and $\epsilon$ is a constant~\cite{ioffe2015batch}.
Typically, following a fully connected layer the batch normalization is computed per activation.
Consider a fully connected layer that performs the matrix multiplication and bias add shown in Equation~\ref{eq:bn_fc}:

\begin{equation}
y_{j} = \sum_{k=1}^{p} x_{k} \cdot w_{kj} + b_{j}
\label{eq:bn_fc}
\end{equation}

\let\vec\mathbf
When the fully connected layer is followed by a batch normalization layer, $\vec{z} = \mathrm{BN}(\vec{x}W+\vec{b})$, the batch normalization can be folded into the weights and biases of the fully connected layer, as show in Equation~\ref{eq:bn_fold}:

\begin{equation}
z_{j} = \sum_{k=1}^{p}
x_{k} \cdot 
\underbrace{ c_j \cdot w_{kj}  }_{w_{kj}'} + \underbrace{c_j \cdot b_{j} + d_j}_{b_{j}'} 
\label{eq:bn_fold}
\end{equation}

resulting in a fully connected layer performing the operation $\vec{z} = \vec{x} W' + \vec{b}' $.
Since convolutions can be mapped to fully connected layers, and batch normalization in CNNs is per channel, we can apply the same optimization.

\section{Novel activation functions}\label{sec:asymmetric}

\newcommand{\f}[1]{\mathrm{#1}}

Two more recently developed activation functions are Swish (Equation~\ref{eq:swish})~\cite{ramachandran2017searching} and GELU (Equation~\ref{eq:gelu})~\cite{hendrycks2016gaussian}, used in EfficientNets and BERT, respectively.

\begin{equation}
\f{swish}(x)=x\cdot\f{sigmoid}(x)
\label{eq:swish}
\end{equation}

\begin{equation}
\f{GELU}(x)=\frac{x}{2}(1+\f{erf}(\frac{x}{\sqrt{2}}))
\label{eq:gelu}
\end{equation}

\begin{figure}[hb]
\centering
\begin{subfigure}{0.49\textwidth}
\includegraphics[width=\textwidth, clip=0 0 10 0]{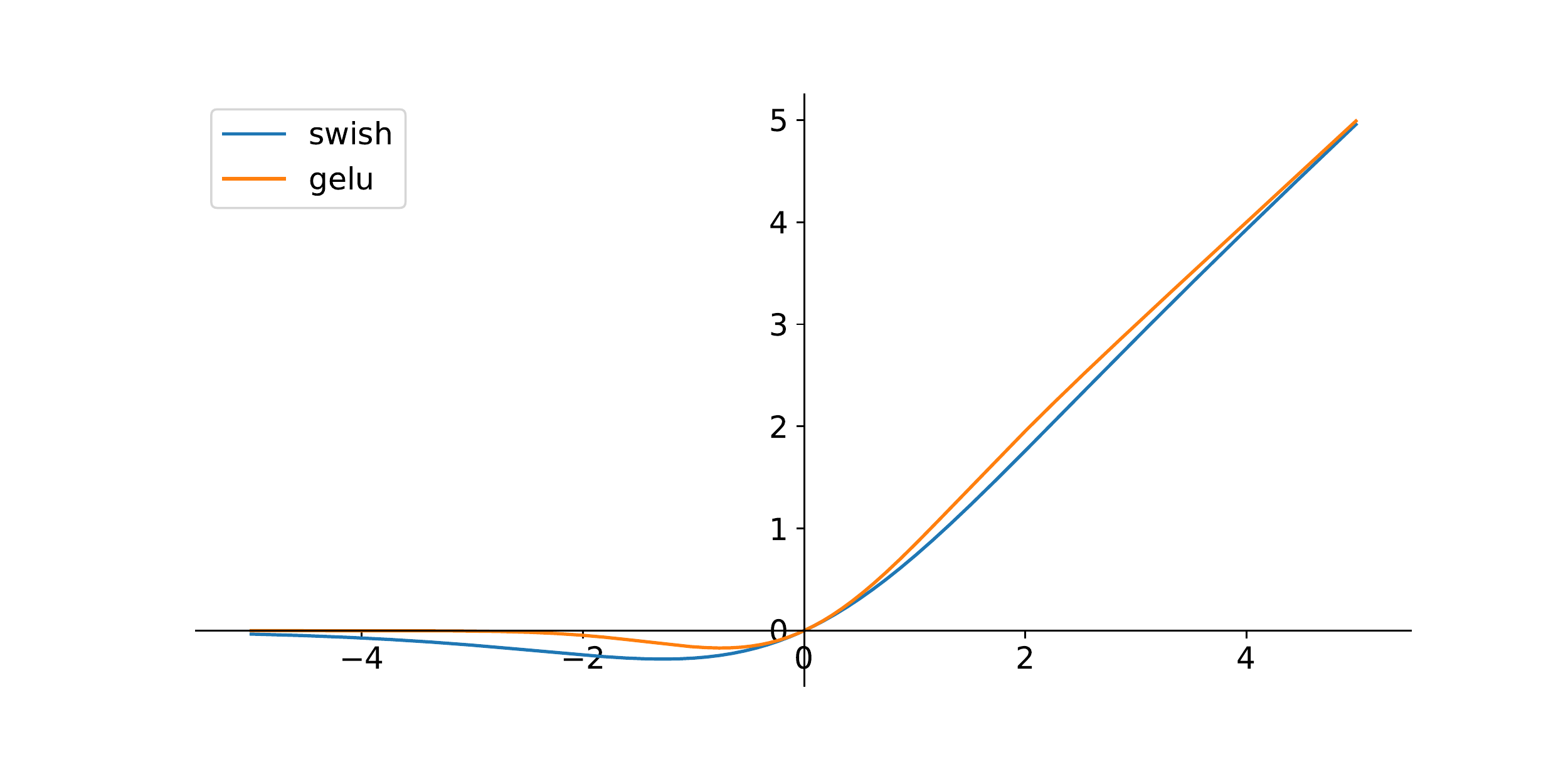}
\caption{GELU and Swish}
\label{fig:gelu_swish}
\end{subfigure}%
\begin{subfigure}{0.49\textwidth}
\includegraphics[width=\textwidth, clip=0 0 10 0]{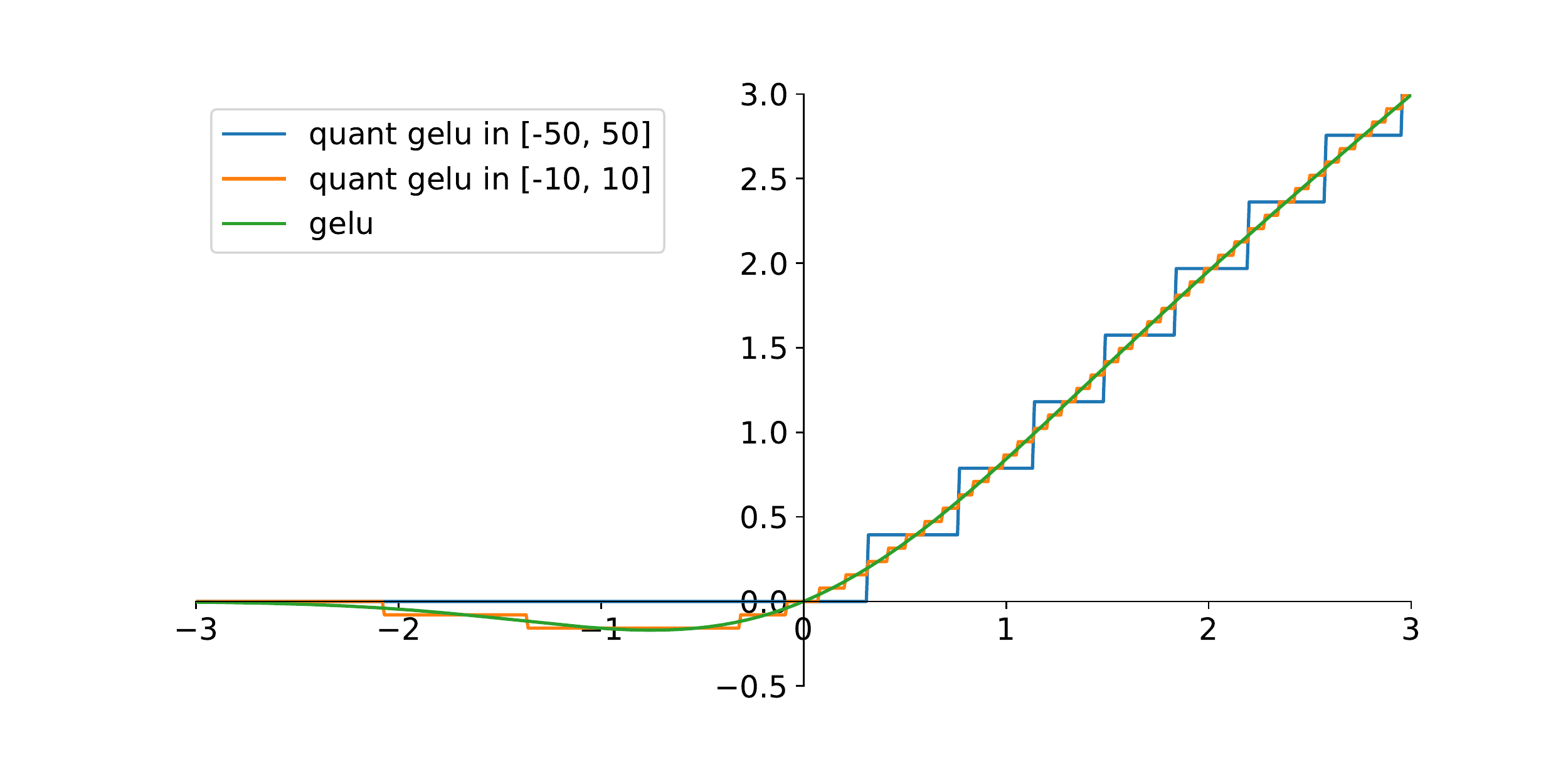}
\caption{Quantizing GELU by scale quantization with different ranges}
\label{fig:quant_gelu}
\end{subfigure}
\caption{Quantization mapping of real values to int8}
\label{fig:gelu}
\end{figure}

These activation functions, shown in Figure~\ref{fig:gelu_swish}, are both smooth and ReLU-like but with small, bounded negative output ranges. 
Specifically, Swish has an output range of $[-0.2785,\infty]$ and GELU has an output range of $[-0.1700, \infty]$.
This poses a challenge for uniform quantization as it should represent both small negative values and large positive values.

Figure~\ref{fig:quant_gelu} shows the composition of GELU and fake quantization with different symmetric ranges.
If the output of GELU is quantized to {[}-50, 50{]}, then all negative values will round to zero. 
However, if we restrict the range to {[}-10, 10{]} then two negative values can be represented. 
Table \ref{table:gelu10} shows the accuracy of post training quantization with GELU outputs clipped to 10 (GELU10), and then calibrated with max calibration.
Just by clipping the output of GELU we can achieve the best post training quantization accuracy with a simple max calibration, exceeding the previous best activation calibration by 0.46 F1. 
Furthermore, this result almost matches the best QAT result of 90.67 F1. 

\begin{table}[hb]
\centering
\begin{tabular}{l|c|cccccc|c}
\toprule
Model & fp32  & Max   & Entropy & 99.9\% & 99.99\% & 99.999\% & 99.9999\% & Max with GELU10 \\
\midrule
BERT Large      & 91.01 & 85.92 & 37.40    & 26.18  & 89.59   & 90.20 & 90.10 & 90.66 \\
\bottomrule
\end{tabular}
\caption{BERT int8 post training quantization. Comparing previous calibration results to max with GELU10.}
\label{table:gelu10}
\end{table}

\end{document}